\documentclass[10pt, a4paper]{article}

\usepackage[final]{lrec2026} %

\usepackage{microtype}
\usepackage{graphicx}
\usepackage{booktabs}
\usepackage{xcolor,colortbl}
\usepackage{hyperref}
\usepackage{xspace}
\usepackage{caption}
\usepackage{varwidth}
\usepackage{multicol}
\usepackage{multirow}
\usepackage{comment}
\usepackage[normalem]{ulem}
\usepackage{makecell}

\usepackage{tikz}
\usetikzlibrary{calc, positioning, shapes, arrows.meta, decorations.pathmorphing}

\definecolor{bluegray}{rgb}{0.4, 0.6, 0.8}

\newcommand{\hmodify}[1]{\textcolor{black}{#1}}

\newcommand{\meanpmstdev}[2]{#1\textcolor{gray}{\small$\pm$#2}}

\newcommand{\project}{{Pantagruel}\xspace}%
\newcommand{\inaCentK}{{{INA-100k}}\xspace}

\newcommand{\camemBWk}{\mbox{{CamemBERT-B-Wk}}\xspace} %
\newcommand{\camemBOsc}{\mbox{{CamemBERT-B-Osc}}\xspace} %
\newcommand{\flauB}{\mbox{{FlauBERT-B}}\xspace} 

\newcommand{\wvB}{\mbox{{LeBenchmark-w2v-B-1k}}\xspace}

\newcommand{\wvL}{\mbox{{LeBenchmark-w2v-L-7k}}\xspace}
\newcommand{\wvLL}{\mbox{{LeBenchmark-w2v-L-14k}}\xspace}

\newcommand{\pantaWk}{\mbox{{Pantagruel-B-camtok-Wk}}\xspace}
\newcommand{\pantaCrs}{\mbox{{Pantagruel-B-Wk}}\xspace}

\newcommand{\pantaWkMLM}{\mbox{{Pantagruel-B-Wk-MLM}}\xspace}
\newcommand{\pantaCrsII}{\mbox{{Pantagruel-B-Crs-MLM}}\xspace}

\newcommand{\pantaOscMLM}{\mbox{{Pantagruel-B-Osc-MLM}}\xspace}
\newcommand{\pantaCrsMLM}{\mbox{{Pantagruel-B-Crs-MLM}}\xspace}

\newcommand{\pantaspeechB}{\mbox{{Pantagruel-B-1k}}\xspace}
\newcommand{\pantaspeechBLB}{\mbox{{Pantagruel-B-14k}}\xspace}
\newcommand{\pantaspeechL}{\mbox{{Pantagruel-L-14k}}\xspace}
\newcommand{\pantaspeechLL}{\mbox{{Pantagruel-L-114k}}\xspace}

\newcommand{\panta}{\mbox{\project}\xspace}

\makeatletter
\DeclareRobustCommand\onedot{\futurelet\@let@token\@onedot}
\def\@onedot{\ifx\@let@token.\else.\null\fi\xspace}
\def\eg{\emph{e.g}\onedot} 
\def\ie{\emph{i.e}\onedot} 
 
\def\etc{\emph{etc}\onedot}

\makeatother

\title{Pantagruel: Unified Self-Supervised Encoders \\ for French Text and Speech
}

\name{
Phuong-Hang Le$^{1,9}$,
Valentin Pelloin$^{2}$,
Arnault Chatelain$^{4}$,
Maryem Bouziane$^{3}$,\\\large \bfseries
Mohammed Ghennai$^{1}$,
Qianwen Guan$^{5}$,
Kirill Milintsevich$^{2}$,\\\large \bfseries
Salima Mdhaffar$^{3}$,
Aidan Mannion$^{1}$,
Nils Defauw$^{6}$,
Shuyue Gu$^{5}$,\\\large \bfseries
Alexandre Audibert$^{1}$,
Marco Dinarelli$^{1}$,
Yannick Estève$^{3}$,
Lorraine Goeuriot$^{1}$,\\\large \bfseries
Steffen Lalande$^{2}$,
Nicolas Hervé$^{2}$,
Maximin Coavoux$^{1}$,
François Portet$^{1}$,\\\large \bfseries
Étienne Ollion$^{4}$,
Marie Candito$^{5}$,
Maxime Peyrard$^{1}$,
Solange Rossato$^{1}$,\\\large \bfseries
Benjamin Lecouteux$^{1}$,
Aurélie Nardy$^{7}$,
Gilles Sérasset$^{1}$,
Vincent Segonne$^{8}$,\\\large \bfseries
Solène Evain$^{10}$,
Diandra Fabre$^{1}$,
Didier Schwab$^{1}$
}

\address{
$^{1}$ Univ. Grenoble Alpes, CNRS, Grenoble INP, LIG, 38000 Grenoble, France \\
$^{2}$ INA (Institut National de l'Audiovisuel), 4 Avenue de l'Europe, 94366 Bry-sur-Marne, France \\
$^{3}$ Avignon Université, LIA, France \\
$^{4}$ CREST (École Polytechnique, ENSAE, CNRS), 5 avenue Le Chatelier, 91120 Palaiseau, France \\
$^{5}$ LLF (Université Paris Cité and CNRS), UFRL Olympe de Gouges, \\13 place Paul Ricoeur, 75013 Paris, France \\
$^{6}$ Univ. Grenoble Alpes, EFELIA-MIAI, IUT2 Grenoble, LIG, 38000 Grenoble, France \\
$^{7}$ Univ. Grenoble Alpes, Lidilem, 38000 Grenoble, France \\
$^{8}$ Université Bretagne Sud, CNRS, IRISA, France \\
$^{9}$ Saclay AI, France\\
$^{10}$ IRIT, Université de Toulouse, CNRS, Toulouse INP, UT3, Toulouse, France 
}

\abstract{
We release \project models, a new family of self-supervised encoder models for French text and speech. Instead of predicting modality-tailored targets such as textual tokens or speech units, \project learns contextualized target representations in the feature space, allowing modality-specific encoders to capture linguistic and acoustic regularities more effectively. Separate models are pre-trained on large-scale French corpora, including Wikipedia, OSCAR and CroissantLLM for text, together with MultilingualLibriSpeech, LeBenchmark, and \inaCentK for speech. \inaCentK is a newly introduced 100\,000-hours corpus of French audio derived from the archives of the \textit{Institut National de l’Audiovisuel} (INA), the national repository of French radio and television broadcasts, providing highly diverse audio data. We evaluate \project across a broad range of downstream tasks spanning both modalities, including those from the standard French benchmarks such as FLUE or LeBenchmark. Across these tasks, \project models show competitive or superior performance compared to strong French baselines such as CamemBERT, FlauBERT, and LeBenchmark 2.0, while maintaining a shared architecture that can seamlessly handle either speech or text inputs. These results confirm the effectiveness of feature-space self-supervised objectives for French representation learning and highlight \project as a robust foundation for multimodal speech–text understanding.
\\ %
\Keywords{self-supervised, data2vec, JEPA, French, speech, text, representation learning, predictive modeling}
}

\begin{document}

\maketitleabstract

\section{Introduction}

Mirroring trends in other languages, self-supervised encoders have become the standard backbone for French speech and language processing. Text models such as FlauBERT~\citep{le2020flaubertunsupervisedlanguagemodel} and CamemBERT~\cite{martin-etal-2020-camembert}, together with LeBenchmark~\cite{evain:hal-03407172,parcollet:hal-04441389} for speech, have established strong baselines across a variety of downstream tasks. Yet most prior work relies on token-level reconstruction~\cite{devlin2019bert, Baevski2020wav2vec2, Warner2025Smarter}, which can under-utilize the structural regularities of continuous signals and hinders a unified treatment of text and speech. Recent predictive approaches~\cite{lecun2022path,Assran2023Selfsupervised,Baevski2023Efficient} that learn contextualized targets in feature space offer a promising alternative, enabling modality-specific encoders to capture richer linguistic and acoustic structure beyond the surface form.

In this paper, we introduce \textbf{\project}, a family of French self-supervised encoders for text and speech, trained separately using the data2vec 2.0 architecture \citep{Baevski2022data2vec}, an instance of the Joint Embedding Predictive Architecture~(JEPA) framework~\cite{lecun2022path}. \project follows a teacher–student training paradigm in which the student predicts masked latent representations produced by a teacher that observes the full, unmasked input. Such representation-based objectives have proven highly effective for vision~\cite{Assran2023Selfsupervised,mo2024connecting} and audio~\cite{Fei2023AJEPA,tuncay2025audiojepa,yuksel2025wavjepa}, yet they remain underexplored for text. Recent work suggests that textual tokens are compact, semantically dense units with minimal low-level variability, leaving limited room for embedding-based methods to improve over input-level approaches~\cite{van2025joint}. Our experiments on text-based models also confirm this hypothesis. Therefore, for text, we propose to augment the feature-space objective with masked language modeling \cite[MLM,][]{devlin2019bert} to better capture fine-grained syntactic and semantic information, yielding stronger textual representations while retaining the benefits of contextualized target prediction. Our self-supervised models are released on the HuggingFace hub\footnote{\url{https://huggingface.co/PantagrueLLM}}.

To support large-scale pre-training in French, we curate substantial corpora for each modality. For text, we use %
Wikipedia, OSCAR~\cite{martin-etal-2020-camembert} and CroissantLLM~\cite{faysse2024croissantllm} datasets. For speech, we assemble a diverse collection spanning read, spontaneous, professional, and broadcast speech audio data, including 14\,000 audio hours from LeBenchmark~\cite{evain:hal-03407172} and 100\,000 audio hours from \inaCentK, a new corpus derived from the archives of France's National Audiovisual Institute~(INA)
that we introduce 
in this paper.
We evaluate \project on a broad suite of downstream tasks in both modalities and compare it to three main French baselines: FlauBERT \citep{le2020flaubertunsupervisedlanguagemodel} and CamemBERT \citep{martin-etal-2020-camembert} for text, and LeBenchmark~\cite{evain:hal-03317730,evain:hal-03407172,parcollet:hal-04441389} for speech.

\paragraph{Contributions}
First, we release \project, a family of French self-supervised encoders for speech and text based on the data2vec~2.0 and JEPA frameworks, which allows training models on different modalities using the same framework.
Second, we investigate embedding-based prediction objectives for text, an underexplored regime, and show that combining feature-space prediction with MLM yields competitive results for French text encoders.
Third, we study the impact of the large-scale \inaCentK broadcast corpus on the models' performances.
Finally, we provide a unified evaluation across speech and text tasks, where \project consistently matches or improves over established French baselines.

\begin{figure*}[t!]
    \centering
    \includegraphics[width=\textwidth]{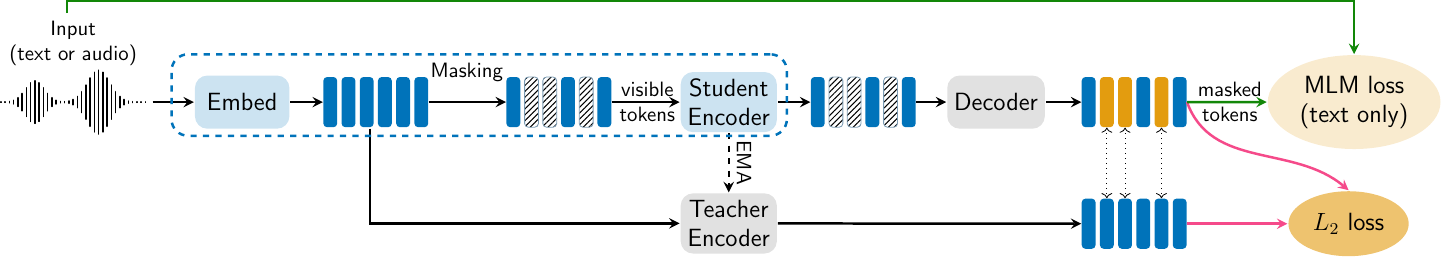}
    \caption{Overview of the \project model architecture. The network starts with a modality-specific pre-net to extract feature vectors from the input text/speech sequence. These features are input to a teacher encoder, while randomly chosen visible tokens (in blue) are input to a student encoder. A lightweight decoder predicts the teacher’s latent representations from the student’s outputs. For text input, an additional masked language modeling~(MLM) loss is used. The teacher’s parameters are updated as an exponential moving average (EMA) of the student's. After training, only the embedding layer and the student encoder are used for fine-tuning on downstream tasks. 
    }
    \label{fig:model_overview}
\end{figure*}

\section{Related Work}

Self-supervised learning (SSL) has driven rapid progress in several domains, in particular in text and speech processing. In text, encoder-only models such as BERT~\cite{devlin2019bert} learn bidirectional representations through the MLM objective, with subsequent refinements improving attention mechanisms, efficiency, and context length~\citep{clark2020electra,he2020deberta,Lan2020,Warner2025Smarter}, while the GPT family~\citep{Radford2018ImprovingLU,radford2019language,brown2020language} popularized autoregressive pre-training for generative tasks. In speech, wav2vec2.0~\citep{Baevski2020wav2vec2} introduced contrastive learning over quantized representations, while HuBERT~\citep{hsu2021hubert} adopted masked prediction of discrete clusters, and WavLM~\citep{chen2022wavlm} improved robustness through denoising objectives. Initially developed for English, these self-supervised frameworks were rapidly extended to multilingual settings~\citep{devlin2019bert,conneau-etal-2020-unsupervised,scao2022bloom,shliazhko2024mgpt,babu2022xls}.

For French, many studies have adapted these self-supervised architectures to both modalities. In text, FlauBERT~\citep{le2020flaubertunsupervisedlanguagemodel} and CamemBERT~\citep{martin-etal-2020-camembert} were the first French BERT variants, trained on large-scale corpora and demonstrating that language-specific pre-training outperforms multilingual models on downstream tasks. Subsequent efforts explored efficiency trade-offs with compact models such as FrALBERT~\citep{cattan2022usability}, LePetit~\citep{michelid2020importance}, and D'AlemBERT~\citep{gabay2022freem}. Encoder-decoder and decoder-only approaches have also been explored with BARThez~\citep{eddine2021barthez}, PAGnol~\citep{launay2022pagnol}, and Cedille~\citep{muller2022cedille}, while more recent large-scale models such as CroissantLLM~\cite{faysse2024croissantllm} further bridge the gap with multilingual systems. Continued refinement of French encoders has led to improved variants including CamemBERTa~\cite{antoun2023data}, CamemBERT~2.0~\cite{antoun2024camembert2}, and CamemBERTav2~\cite{antoun2024camembert2}, which revisit data filtering, tokenization, and training recipes to yield stronger representations. For speech, the LeBenchmark initiative~\citep{evain:hal-03407172,parcollet:hal-04441389} extends the pre-training paradigm to spoken French, providing wav2vec2.0-based models and evaluation resources for speech understanding in French. For spoken language understanding, \mbox{FlauBERT-Oral}~\citep{pelloin22_interspeech} provides textual representations adapted to transcribed spoken French. Altogether, these efforts contribute to a comprehensive ecosystem of French models covering written, specialized, and oral modalities.

Orthogonal to the above advances, an active line of research has explored %
embedding-level predictive objectives, %
where models learn to predict latent representations of masked regions from visible context%
. A notable example is data2vec~\citep{Baevski2022data2vec,Baevski2023Efficient}, which introduced a modality-agnostic framework for predicting contextualized representations across speech, vision, and text. Closely related ideas were formalized in the JEPA framework~\citep{lecun2022path}, which emphasizes prediction in the latent space. %
This paradigm has since enabled successful applications across domains: I-JEPA~\citep{Assran2023Selfsupervised}, V-JEPA~\citep{Bardes2024Revisiting,Assran2025vjepa} for image and video; A-JEPA~\citep{Fei2023AJEPA}, Audio-JEPA~\citep{tuncay2025audiojepa}, and WavJEPA~\citep{yuksel2025wavjepa} for audio%
. Similar to data2vec, most of these work rely on the teacher-student setup to prevent representation collapse. Recently, LeJEPA~\cite{Balestriero2025lejepa} introduced a theoretically grounded formulation based on Latent-Euclidean regularization, removing the need for such heuristics. We leave its exploration to future work. Beyond data2vec, JEPA-style pre-training remains largely underexplored for text, where the input-level MLM objective continues to dominate. In this work, we investigate JEPA-style representation learning for French speech and text using data2vec2.0 architecture~\citep{Baevski2023Efficient}. For speech, we adopt a pure feature-based objective. For text, we propose a hybrid approach that combines JEPA-style representation prediction with MLM to leverage their complementary strengths, inspired by \citep{huang2025llm}. %

\section{Models and Pre-training Framework}
\label{sec:method}

Our SSL models for speech are entirely based on data2vec~2.0 \cite{Baevski2023Efficient}. For text, we extended the data2vec~2.0 loss to include the MLM objective~\cite{devlin2019bert,liu2019roberta}, which we found to produce better textual representations. Training is performed separately for each modality, following the data2vec approach.

\subsection{Framework overview}

An overview of our pre-training framework is presented in Figure~\ref{fig:model_overview}. Given an input audio or a sequence of text tokens, a modality-specific \textbf{pre-net} is used to extract a sequence of feature vectors. For speech, this pre-net is typically a small CNN while for text it is a standard embedding layer. Then, the obtained sequence is fed into a \textbf{teacher encoder}. At the same time, randomly selected items
(called \emph{visible tokens}) of this sequence are fed into the \textbf{student encoder} (the unselected ones are called \emph{masked tokens}). A lightweight, modality-specific %
convolutional \textbf{decoder} predicts the teacher’s latent representations from the student’s encoder outputs and is used only in pre-training. Both teacher and student encoders share the same Transformer encoder architecture~\cite{Vaswani2017Attention}. The teacher’s parameters are maintained as an exponential moving average (EMA) of the student’s, gradually stabilizing during training. Conceptually, this approach can be viewed as an instance of the predictive modeling paradigm~\cite{lecun2022path,Assran2023Selfsupervised}, sharing its principle of learning by predicting latent representations rather than reconstructing raw inputs. The idea of using \emph{masked feature prediction} within a \emph{self-distillation} setup is applicable across diverse modalities, motivating our exploration of this framework as a step toward building a truly multi-modal encoder.

\paragraph{Masked feature prediction}\label{sec:method-loss} The training loss is defined as the $L_2$ distance between the student predictions (\ie, the outputs of the decoder), and the teacher encoder's representations, computed over the masked regions. The teacher representations are not obtained from the last Transformer layer but by averaging the last $K$ layers, for better contextualization. %
To improve efficiency, multiple masked versions of each example share a single teacher forward pass. %

\paragraph{Extension for better textual representations} While embedding-based objective excels in speech and vision domains \cite{Assran2023Selfsupervised,Fei2023AJEPA,Bardes2024Revisiting,Assran2025vjepa}, token-level objectives remain dominant in text modeling. Unlike continuous signals, text is discrete and sparse, thus predicting global contextual embeddings alone may not adequately capture fine-grained syntactic and semantic information essential for downstream tasks \cite{Mo2024DMTJEPA}. Combining masked feature prediction with token-level objectives, as explored in recent work \cite{huang2025llm}, offers a promising direction. Following this idea, we enhance our non-generative text models by combining the MLM objective with the data2vec loss, encouraging the model to capture local semantics alongside rich contextual representations. We find that using the student decoder to predict masked text tokens (see Figure~\ref{fig:model_overview}) outperformed the BERT-style implementation, which requires a second forward pass in our framework. We thus adopt the decoder-based approach for our text models.

\subsection{\project model configurations and implementation details}\label{sec:model-configurations-and-implementation-details}

\paragraph{Tokenizers} We train our tokenizer on a subset of the CroissantLLM dataset \cite{faysse2024croissantllm} with customized settings tailored for French, such as normalizing apostrophe variants and treating them as single tokens in elided words %
(\eg, \textit{quelqu'}, \textit{c'}). For comparison, we also experimented with the %
CamemBERT tokenizer \cite[][denoted \emph{camtok}]{martin-etal-2020-camembert}. While CamemBERT followed the original BERT~\cite{devlin2019bert} and used a vocabulary size of 32K tokens, we used a larger size of 50K, following more recent works such as GPT-2~\cite{radford2019language}, RoBERTa~\cite{liu2019roberta}, and ModernBERT~\cite{Warner2025Smarter}. Later in Section~\ref{sec:ablation-study}, we include two more tokenizers that we train with different settings to investigate the effect of the tokenizer on our models.

\paragraph{Model configurations} Following standard Transformer architectures, we propose two model configurations: \textbf{base} and \textbf{large}. The base variant has 12 layers, 8 attention heads, and 768 hidden dimensions. For the large variant, these values are, respectively, 24, 16, and 1024. A summary of our models is given in Table~\ref{tab:model_details}. For text models, we only include the base variant as we experienced training instability with the large variant. It should be noted that the data2vec papers~\cite{Baevski2022data2vec,Baevski2023Efficient} also did not report results for the large variant on text tasks. Investigating large text models is left for future work.

\begin{table}[!htb]
    \centering
    \resizebox{\linewidth}{!}{
    \begin{tabular}{llrl}
    \toprule
     & \textbf{Model name} & \textbf{Param} & \textbf{Trained on} \\
    \midrule
    \multirow{5}{*}{\rotatebox{90}{Text}} & \pantaWk & 110M  & Wikipedia\\
    & \pantaCrs & 125M  & Wikipedia\\
    & \pantaWkMLM & 125M  & Wikipedia\\
    & \pantaOscMLM & 125M  & OSCAR \\
    & \pantaCrsMLM &125M  & CroissantLLM\\
    \midrule
    \multirow{4}{*}{\rotatebox{90}{Speech}} & \pantaspeechB & 93M & LibriSpeech\\
    & \pantaspeechBLB & 93M & LeBenchmark\\
    & \pantaspeechL & 313M & LeBenchmark \\
    & \pantaspeechLL & 313M & LeBenchmark + INA\\
    \bottomrule
    \end{tabular}
    }
    \caption{\hmodify{\project model names and configurations. Suffixes: ``B/L'' mean \emph{base}/\emph{large} architecture, and ``camtok'' denotes the CamemBERT tokenizer. For text, the third suffix specifies the dataset name, while for speech it indicates the training dataset size in hours (\eg, ``1k'' means 1\,000 audio hours).}}
    \label{tab:model_details}
\end{table}

\paragraph{Loss functions} Our speech models follow data2vec~2.0 and use only masked feature prediction, $\mathcal{L}_{\textrm{speech}} = \mathcal{L}_{L_2}$ (see Section~\ref{sec:method-loss}). For text, we propose a hybrid loss $\mathcal{L}_{\textrm{text}} = \mathcal{L}_{L_2} + \lambda \mathcal{L}_{\textrm{mlm}}$, where $\lambda$ decays linearly from $\lambda_{\textrm{start}}$ to $\lambda_{\textrm{end}}$ over $N_{\lambda}$ training steps and remains fixed thereafter. We find this schedule yields more stable training and better performance than a constant weight, similar to prior work showing that auxiliary objectives are most beneficial early but can interfere with the main objective later~\cite{du2018adapting,fu2019cyclical}.

\paragraph{Implementation details}
Following data2vec~2.0 recipe, we used the Adam optimizer \cite{kingma2015adam} with cosine learning rate scheduler \cite{Loshchilov2017SGDR}, where the learning rate $\eta$ is increased linearly from $\eta_{\min}$ to the configured maximum learning rate $\eta_{\max}$ during a warmup %
of $N_{\textrm{warmup}}$ steps, then decreases smoothly following a cosine curve until training reaches the specified maximum number of updates $N_{\textrm{max}}$ where $\eta = \eta_{\min}$. Details of the pre-training hyperparameters are provided in Table~\ref{tab:speech-pretraining-hyperparams} and~\ref{tab:text-pretraining-hyperparams} for speech and text models, respectively. %

\section{Datasets and Resources}

This section describes data collection and organization for French text and speech. %
Our goal was to maximize dataset size, including both aligned (audio with transcripts) and unaligned data. In the following, we detail the publicly available datasets, the proprietary INA corpus, and preprocessing steps.

\subsection{Text datasets}

Table~\ref{tab:text_datasets} summarizes the data used to train our text models. We leveraged monolingual French data from three public sources: the 2019 Wikipedia dump from FlauBERT~\cite{le2020flaubertunsupervisedlanguagemodel} (876M tokens), the French portion of OSCAR~\cite{OrtizSuarezSagotRomary2019} (32.7B tokens), and the French subset of the CroissantLLM dataset~\cite{faysse2024croissantllm}, which is composed of web data (292B tokens), legal and administrative (5.3B tokens), cultural (2.7B tokens), encyclopedia (2B tokens), and industrial data (191M tokens). The corresponding models trained on these datasets are listed in Table~\ref{tab:model_details}.

\begin{table}[!hbt]
\centering
\resizebox{\columnwidth}{!}{
\begin{tabular}{l r r l}
\toprule
\textbf{Dataset} & \textbf{Tokens} & \textbf{Size} & \textbf{Type} \\
\midrule
Wikipedia (Jul 2019) &   876M &   4 GB & Encyclopedia \\
OSCAR (Nov 2018)     &  32.7B & 138 GB & Web crawl \\
CroissantLLM         & 302.2B & 1.3 TB & Mix \\
\bottomrule
\end{tabular}
}
\caption{Statistics of pre-training text corpora.}
\label{tab:text_datasets}
\end{table}
\subsection{Speech datasets}

The complete list of datasets considered in this study for speech and their statistics are summarized in Appendix~\ref{sec:speech_dataset_appendix}. %
We used the dataset collection of LeBenchmark for French which gathered a wide variety of French speech corpora covering different accents, acted emotions, telephone dialogues, read speech, spontaneous sentences, and professional speech. This collection includes Multilingual LibriSpeech (MLS) dataset \cite{old_pratap_mls_2020}. 
It is worth noting that some of these corpora are transcribed either manually or automatically and are therefore representative of aligned speech/text data. %
We refer the reader to \citet{parcollet:hal-04441389} for the details.

However, these remain far from SOTA models, such as Whisper (\citealp{radford2022robustspeechrecognitionlargescale}, trained on 680k hours of labeled data) or Google USM~(\citealp{zhang2023googleusmscalingautomatic}, trained on more than 12 million hours of multilingual data). Thus, to increase the pre-training speech dataset, a corpus of 100\,000 hours of audio content is obtained in partnership with INA, the French National Audiovisual Archive which we will refer to as \inaCentK in the rest of the paper. In France, INA is in charge of collecting and archiving TV and radio broadcast since 1975. We evaluate that 75\% of this audio content is speech. %

The pre-processing steps used to prepare \inaCentK are described below.
An initial corpus of 473k hours of content, broadcast on 113 French TV and Radio channels, from 1940 to 2022 was collected, covering various kinds of audiovisual content: news, adverts, documentaries, game shows, movies, musics, cartoons, sports, etc.
        
We first applied the audio deduplication tool presented by \citet{chenot:hal-01017118} to remove duplicate content. This tool extracts audio fingerprints. When at least four consecutive similar fingerprints are detected, all other matches are discarded. Deduplication helps control distribution and reduce bias \citep{lee2022deduplicating}.
Multiple standard benchmark evaluation corpora are also based on French TV broadcast content: we used the same tool to remove them from the INA corpus. This include, among others,
ESTER1 \cite{gravier-etal-2004-ester},
ESTER2 \cite{galliano09_interspeech},
EPAC \cite{esteve-etal-2010-epac},
QUAERO \cite{boudahmane-etal-2011-advances},
ETAPE \cite{gravier-etal-2012-etape},
REPERE \cite{giraudel-etal-2012-repere}.
Overall, these two deduplication steps removed 154k hours (33\%) from the initial corpus. 
Finally we randomly sampled 12M audio chunks of 30s to create a 100\,000 hours training corpus, denoted as \inaCentK.
        
\section{Ablation study}\label{sec:ablation-study}
In this section, we investigate the effect of different tokenizers on our text models and the impact of varying pre-training data sizes on our speech models. The ablation results are reported on the validation sets of two representative tasks: natural language inference (XNLI dataset) for text and automatic speech recognition (CommonVoice dataset) for speech. Further details on these datasets are provided in Sections~\ref{sec:text-models-evaluation} and~\ref{sec:speech-models-evaluation}.

\paragraph{Effects of different text tokenizers}
In addition to our main ``customized'' tokenizer (used in Section~\ref{sec:benchmarks}) and the CamemBERT one (see Section~\ref{sec:model-configurations-and-implementation-details}), we trained two other tokenizers, ``default'' and ``basic'', using HuggingFace's \textit{tokenizers} library. %
We refer to the documentation of this library for the details of these settings. The ablation study is conducted on the Wikipedia dataset with models trained for 500K steps (approximately 10 epochs over the dataset). As shown in Table~\ref{tab:ablation_tokenizers}, our main tokenizer outperforms the others on the XNLI dataset.

\begin{table}[!htb]
    \centering
    \begin{tabular}{ll}
    \toprule
    { \textbf{Tokenizers}} & \textbf{Accuracy}\\
    \midrule
    {Camtok} &  \meanpmstdev{76.94}{0.51} \\
    Default & \meanpmstdev{76.34}{0.52} \\
    Basic & \meanpmstdev{77.02}{0.55} \\
    Customized & \meanpmstdev{\textbf{77.40}}{0.46} \\
    \bottomrule
    \end{tabular}
    \caption{Results using different tokenizers on the dev sets of the XNLI dataset (average over 10 runs). ``Customized'' denotes our main tokenizer.%
    }
    \label{tab:ablation_tokenizers}
\end{table}

\paragraph{Effects of pre-training data size for speech}
We trained four speech models: two models using the base architecture on 1K hours of LibriSpeech and 14K hours of the LeBenchmark dataset, and two models using the large architecture on 14K hours of LeBenchmark and 114K hours, which combines LeBenchmark 14K hours with 100K hours from the INA dataset. Table~\ref{tab:ablation_speech_data} reports the Word Error Rate (WER) on the CommonVoice dev sets. Increasing the data from 1K to 14K hours for the base architecture yields a modest improvement in WER (from 8.92 to 8.46). For the large architecture, however, increasing the data from 14K to 114K hours slightly hurts the results. We hypothesize that this could be due to the data nature of INA being quite different from CommonVoice, while at the same time the model capacity being potentially too limited to sufficiently fit this large dataset.

\begin{table}[t]
    \centering
    \begin{tabular}{l | cccc}
    \toprule
    \multirow{2}{*}{ \textbf{Model/Data}} & \multicolumn{2}{c}{\textbf{Base}} & \multicolumn{2}{c}{\textbf{Large}} \\
    \cmidrule(lr){2-3} \cmidrule(lr){4-5}
     & \textbf{1K} & \textbf{14K} & \textbf{14K} & \textbf{114K} \\
    \midrule
    WER & 8.92 & 8.46 & 6.95 & 7.07 \\
    \bottomrule
    \end{tabular}
    \caption{WER on CommonVoice dev sets across model architectures and data sizes.
    }%
    \label{tab:ablation_speech_data}
\end{table}

\section{Benchmarks}
\label{sec:benchmarks}
This section presents text and speech downstream tasks used for evaluation. %
Unless otherwise noted, all results are averaged over five runs with different random seeds. We use $\uparrow$ (respectively $\downarrow$) to indicate that higher (respectively lower) values correspond to better performance.

An overview of the benchmarks, including task descriptions and evaluatuon metrics, is provided in Table~\ref{tab:overview_text_benchmarks} for text-based tasks and Table~\ref{tab:overview_speech_benchmarks} for speech-based tasks. Further details are provided in Appendix \ref{sec:appendix_bench}.

\subsection{Text models evaluation}\label{sec:text-models-evaluation}
\paragraph{Named Entity Recognition (NER)}
We perform NER on PxCorpus \cite{PxCorpus}, a dataset containing 1\,981 French medical prescriptions in the form of speech recordings along with their transcriptions. Each word of the prescription is tagged in the IOB (Inside-Outside-Beginning) format in various chunks such as ``drug name'', ``frequency'', ``dosage'', \etc. In the case of the text modality, the task involves classifying tokens, aligning them with the corresponding words, and assigning an IOB label to each word by taking the label of the first token of the word. 
The metric used for evaluating the models is the F1-score from the SeqEval library \cite{seqeval} which is designed for IOB-tagging evaluation.

\paragraph{Automatic Coreference Resolution (CR)}
The task consists of identifying all mentions (\ie, phrases referring to entities) in the text and grouping those that co-refer to the same entity into clusters. Due to the inherent complexity of coreference resolution, recent systems adopt diverse approaches, often tailored to specific datasets or domains. 
For benchmarking, we selected the WL-Coref model \citep{dobrovolskii-2021-word,doosterlinck-etal-2023-caw,liu-etal-2024-mscaw}, known for its strong performance and computational efficiency during training and evaluation. All experiments were implemented using the Stanza library \citep{qi2020stanza}. 
For evaluation, we used ANCOR, a corpus comprising manually transcribed interviews \citep{muzerelle-etal-2014-ancor}. %
System performance was assessed using the official \texttt{corefud-scorer} \citep{novak-etal-2024-findings}, with head mention matching as the evaluation criterion.
We report the CoNLL F1 score. %

\paragraph{Extractive Question Answering (QA)}
The task consists in finding the shortest span of words in a context answering a given question. We used PIAF1.2 \citep{keraron-EtAl:2020:LREC}, an extractive question answering dataset for French inspired by SQuAD1.1 \citep{rajpurkar2016squad100000questionsmachine} which contains 9\,224 context/question/answer pairs extracted from French Wikipedia. 

The metric used for evaluation is a F1-score on words, where words are categorized in True/False Positives/Negatives depending on their inclusion in the predicted and ground-truth word spans. This F1-score is at the word-level and not the token-level as different models use different tokenizers that segment the same text into varying token counts. As a result using a token-level F1-score would lead to different evaluations for different models.

\paragraph{FLUE benchmark} 
We evaluate the text models on part of the tasks included in the FLUE benchmark  \citep{le2020flaubertunsupervisedlanguagemodel}. The tasks are as follows: text classification, paraphrase identification, natural language inference (NLI), verb sense disambiguation (VSD), and dependency parsing. The overall methodology and downstream model architecture are based on the FLUE framework described by \citet{le2020flaubertunsupervisedlanguagemodel}. For all the above tasks, we reuse the same datasets as in FLUE, except for dependency parsing, for which we use a set of treebanks from the Universal Dependencies repository \citep{de-marneffe-etal-2021-universal}: Sequoia \citep{candito-etal-2014-deep}, GSD \citep{mcdonald-etal-2013-universal}, as well as two treebanks for spoken French: Rhapsodie \citep{lacheret-etal-2014-rhapsodie} and ParisStories \cite{kahane-etal-2021-annotation}.
We use the HOPS parser \citep{grobol:hal-03223424} and report classic evaluation metrics for the task: Labelled Attachment Score (LAS) and part-of-speech accuracy (POS).

\begin{table*}[ht!]
    \centering
    \setlength{\tabcolsep}{4pt} %
    \renewcommand{\arraystretch}{1.0} %
    \resizebox{1.0\textwidth}{!}{
    \begin{tabular}{l | c c c c | c c c c c}
    \toprule
     & \multicolumn{4}{c|}{\textbf{FLUE}} & \textbf{MEDIA} & \textbf{CoNLL} & \textbf{PxCorpus} & \textbf{PIAF} \\
    \cmidrule(lr){2-5}
    \textbf{Model} &
    \multicolumn{1}{c}{\makecell{\textbf{Class.}\\(F1$\uparrow$)}} &
    \multicolumn{1}{c}{\makecell{\textbf{Paraph.}\\(F1$\uparrow$)}} &
    \multicolumn{1}{c}{\makecell{\textbf{NLI}\\(Acc$\uparrow$)}} &
    \multicolumn{1}{c|}{\makecell{\textbf{VSD}\\(F1$\uparrow$)}} &
    \multicolumn{1}{c}{\makecell{\textbf{SLU}\\(CER$\downarrow$)}} &
    \multicolumn{1}{c}{\makecell{\textbf{CR}\\(F1$\uparrow$)}} &
    \multicolumn{1}{c}{\makecell{\textbf{NER}\\(F1$\uparrow$)}} &
    \multicolumn{1}{c}{\makecell{\textbf{QA}\\(F1$\uparrow$)}} \\
    \midrule
    
    \camemBWk   & \meanpmstdev{86.8}{1.2}  & \meanpmstdev{90.7}{0.3} & \meanpmstdev{76.88}{0.3} & 49.44 & \meanpmstdev{10.5}{0.6} & \meanpmstdev{69.2}{0.8} & \meanpmstdev{92.0}{0.2} & \meanpmstdev{47.1}{0.3} \\
    \camemBOsc  & \meanpmstdev{93.7}{0.4}  & \meanpmstdev{91.2}{0.5} & \meanpmstdev{\textbf{82.09}}{0.6} & \textbf{50.03} & \meanpmstdev{11.8}{1.9} & \meanpmstdev{\textbf{73.3}}{0.8} & \meanpmstdev{\textbf{93.6}}{0.2} & \meanpmstdev{\textbf{53.1}}{0.6} \\
    \flauB      & \meanpmstdev{93.3}{0.8}  & \meanpmstdev{89.0}{0.1} & \meanpmstdev{80.60}{N/A}  & 43.93 & \meanpmstdev{10.2}{0.6} & \meanpmstdev{69.8}{0.8} & \meanpmstdev{87.3}{1.6} & N/A \\
    \midrule
    
    \pantaWk    & \meanpmstdev{86.9}{1.1}  & \meanpmstdev{89.0}{0.5} & \meanpmstdev{77.43}{0.6} & 31.83 & \meanpmstdev{12.8}{2.5} & \meanpmstdev{67.9}{0.5} & \meanpmstdev{86.5}{0.8} & \meanpmstdev{43.5}{0.8} \\
    \pantaCrs   & \meanpmstdev{87.3}{0.6}  & \meanpmstdev{89.6}{0.8} & \meanpmstdev{77.90}{0.5} & 29.39 & \meanpmstdev{10.6}{0.6} & \meanpmstdev{65.4}{0.5} & \meanpmstdev{87.0}{1.7} & \meanpmstdev{45.8}{0.4} \\
    \pantaWkMLM   & \meanpmstdev{88.9}{0.7}  & \meanpmstdev{90.0}{0.5} & \meanpmstdev{78.41}{0.5} & 43.28 & \meanpmstdev{10.5}{0.5} & \meanpmstdev{67.8}{0.8} & \meanpmstdev{84.7}{2.5} & \meanpmstdev{47.3}{0.7} \\
    \pantaOscMLM & \meanpmstdev{\textbf{94.0}}{0.3}  & \meanpmstdev{90.9}{0.3} & \meanpmstdev{81.50}{0.5} & 47.90 & \meanpmstdev{10.5}{0.6} & \meanpmstdev{72.6}{0.7} & \meanpmstdev{89.7}{0.4} & \meanpmstdev{51.9}{0.5} \\
    \pantaCrsMLM & \meanpmstdev{93.1}{0.3}  & \meanpmstdev{\textbf{91.6}}{0.4} & \meanpmstdev{81.10}{0.3} & 42.30 & \meanpmstdev{\textbf{10.1}}{0.6} & \meanpmstdev{72.1}{0.7} & \meanpmstdev{86.9}{1.5} & \meanpmstdev{52.9}{0.6} \\
    \bottomrule
    
    \end{tabular}
    }
    \caption{Text results. Evaluation tasks and metrics: {Class.} = text classification (F1); {Paraph.} = paraphrase identification (F1); {VSD} = Verb Sense Disambiguation (F1); {SLU} = Spoken Language Understanding (CER - Concept Error Rate); {CR} = CoNLL (F1); NER = seqeval F1 score evaluated on the PxCorpus dataset; and QA = word-level F1 for extractive question answering evaluated on the PIAF1.2 dataset. FlauBERT was not tested for QA because its tokenizer implementation does not permit word-token alignment.}%

\label{tab:text_eval}
\end{table*}

\begin{table*}[htbp]
\centering
\resizebox{1.0\textwidth}{!}{
\begin{tabular}{l|cccc}
\toprule
\multirow{2}{*}{\textbf{Model}}  &  \textbf{Sequoia}  &  \textbf{GSD}  &  \textbf{Rhapsodie}  &  \textbf{ParisStories} \\
& \textbf{POS/LAS} & \textbf{POS/LAS} & \textbf{POS/LAS} & \textbf{POS/LAS} \\
\midrule
\camemBWk   &   \meanpmstdev{99.1}{0.1} / \meanpmstdev{95.0}{0.1}   &   \meanpmstdev{98.4}{0.1} / \meanpmstdev{95.0}{0.1}   &   \meanpmstdev{97.5}{0.1} / \meanpmstdev{84.6}{0.5}   &   \meanpmstdev{97.1}{0.1} / \meanpmstdev{79.2}{0.3}   \\
\camemBOsc   &   \meanpmstdev{99.0}{0.1} / \meanpmstdev{95.6}{0.1}   &   \meanpmstdev{98.4}{0.1} / \meanpmstdev{95.7}{0.1}   &   \meanpmstdev{97.3}{0.2} / \meanpmstdev{86.2}{0.2}   &   \meanpmstdev{96.8}{0.2} / \meanpmstdev{80.1}{0.3}   \\
\flauB   &   \meanpmstdev{\textbf{99.4}}{0.0} / \meanpmstdev{\textbf{96.0}}{0.0}   &   \meanpmstdev{\textbf{98.7}}{0.1} / \meanpmstdev{\textbf{96.0}}{0.1}   &   \meanpmstdev{\textbf{97.9}}{0.1} / \meanpmstdev{\textbf{86.7}}{0.3}   &   \meanpmstdev{\textbf{97.4}}{0.2} / \meanpmstdev{\textbf{80.9}}{0.3}   \\
\midrule

\pantaWk & \meanpmstdev{99.0}{0.1} / \meanpmstdev{94.3}{0.5} & \meanpmstdev{98.3}{0.1} / \meanpmstdev{94.3}{0.2} & \meanpmstdev{97.1}{0.5} / \meanpmstdev{82.9}{1.1} & \meanpmstdev{96.9}{0.1} / \meanpmstdev{78.5}{0.3}   \\
\pantaCrs & \meanpmstdev{98.5}{0.0} / \meanpmstdev{93.0}{0.3} & \meanpmstdev{98.0}{0.1} / \meanpmstdev{93.5}{0.2} & \meanpmstdev{95.8}{0.4} / \meanpmstdev{80.5}{0.5} & \meanpmstdev{96.3}{0.2} / \meanpmstdev{77.8}{0.4}   \\
\pantaWkMLM & \meanpmstdev{99.0}{0.1} / \meanpmstdev{94.4}{0.3} & \meanpmstdev{98.3}{0.1} / \meanpmstdev{94.8}{0.2} & \meanpmstdev{97.4}{0.2} / \meanpmstdev{83.8}{0.7} & \meanpmstdev{96.9}{0.1} / \meanpmstdev{78.5}{0.6}   \\
\pantaOscMLM & \meanpmstdev{99.3}{0.1} / \meanpmstdev{95.2}{0.4} & \meanpmstdev{98.6}{0.1} / \meanpmstdev{95.3}{0.2} & \meanpmstdev{97.8}{0.1} / \meanpmstdev{85.9}{0.3} & \meanpmstdev{97.2}{0.1} / \meanpmstdev{80.2}{0.3}   \\
\pantaCrsMLM & \meanpmstdev{99.2}{0.1} / \meanpmstdev{95.5}{0.1} & \meanpmstdev{98.5}{0.1} / \meanpmstdev{95.4}{0.1} & \meanpmstdev{97.7}{0.1} / \meanpmstdev{85.6}{0.4} & \meanpmstdev{97.1}{0.1} / \meanpmstdev{79.8}{0.3}   \\

\bottomrule
\end{tabular}
}
\caption{ Text results: dependency parsing ($\uparrow$).}
\label{tab:parsing}
\end{table*}

\begin{table*}[htbp]
    \centering
    \setlength{\tabcolsep}{2pt}
    \renewcommand{\arraystretch}{1.0}
    \resizebox{1.0\textwidth}{!}{
    \begin{tabular}{l|cccccccc}
        \toprule
        \textbf{Dataset}$\rightarrow$ & \textbf{ESSAI-POS} & \textbf{CAS-POS} & \textbf{MEDLINE} & \textbf{EMEA} & \textbf{CAS-SG} & \textbf{E3C-NER} & \textbf{FrMMCQA} & \textbf{CLISTER} \\
        \textbf{Metric}$\rightarrow$ & $\textbf{MacF}_1$ & $\textbf{MacF}_1$ & $\textbf{WF}_1$ & $\textbf{WF}_1$ & $\textbf{WF}_1$ & $\textbf{WF}_1$ & \textbf{Hamming} & $\rho$ \\ \midrule
        \camemBOsc  & \meanpmstdev{96.2}{0.4} & \meanpmstdev{96.6}{0.5} & \meanpmstdev{85.1}{0.2} & \meanpmstdev{93.7}{0.2} & \meanpmstdev{73.4}{2.5} & \meanpmstdev{93.6}{0.8} & \meanpmstdev{34.1}{0.7} & \meanpmstdev{\textbf{87.6}}{0.0} \\
        \camemBWk & \meanpmstdev{96.3}{0.2} & \meanpmstdev{96.5}{0.1} & \meanpmstdev{84.3}{0.3} & \meanpmstdev{93.5}{0.1} & \meanpmstdev{73.9}{0.7} & \meanpmstdev{94.0}{0.2} & \meanpmstdev{32.8}{1.0} & \meanpmstdev{84.6}{0.0} \\
        \flauB & \meanpmstdev{66.8}{0.4} & \meanpmstdev{89.1}{0.2} & \meanpmstdev{81.4}{1.0} & \meanpmstdev{79.5}{1.5} & \meanpmstdev{66.4}{0.7} & \meanpmstdev{\textbf{94.1}}{0.1} & \meanpmstdev{35.1}{0.8} & \meanpmstdev{83.1}{0.0} \\
        \midrule
        \pantaWk & \meanpmstdev{95.7}{0.2} & \meanpmstdev{95.9}{0.2} & \meanpmstdev{82.3}{0.3} & \meanpmstdev{91.8}{0.4} & \meanpmstdev{71.3}{1.1} & \meanpmstdev{93.0}{0.1} & \meanpmstdev{32.7}{0.9} & \meanpmstdev{72.2}{0.0} \\
        \pantaCrs & \meanpmstdev{92.8}{0.9} & \meanpmstdev{89.6}{1.4} &  \meanpmstdev{77.2}{0.2}  & \meanpmstdev{89.9}{0.2} & \meanpmstdev{68.3}{2.3} & \meanpmstdev{92.0}{0.2} & \meanpmstdev{\textbf{35.4}}{2.0} & \meanpmstdev{75.8}{0.0} \\  %
        \pantaWkMLM & \meanpmstdev{96.3}{0.1} & \meanpmstdev{96.8}{0.6} & \meanpmstdev{84.2}{0.4} & \meanpmstdev{93.7}{0.1} & \meanpmstdev{75.7}{0.4} & \meanpmstdev{93.6}{0.3} & \meanpmstdev{33.7}{0.9} & \meanpmstdev{76.9}{0.0} \\ %
        \pantaOscMLM  & \meanpmstdev{96.3}{0.1} & \meanpmstdev{97.0}{0.1} & \meanpmstdev{85.5}{0.3} & \meanpmstdev{\textbf{94.2}}{0.1} & \meanpmstdev{\textbf{76.8}}{0.7} & \meanpmstdev{93.5}{0.1} & \meanpmstdev{35.1}{1.1} & \meanpmstdev{84.1}{0.0} \\  %
        \pantaCrsMLM & \meanpmstdev{\textbf{96.5}}{0.1} & \meanpmstdev{\textbf{96.9}}{0.1} &  \meanpmstdev{\textbf{85.9}}{0.1}  & \meanpmstdev{94.0}{0.2} & \meanpmstdev{76.6}{0.4} & \meanpmstdev{93.5}{0.1} & \meanpmstdev{35.0}{0.7} & \meanpmstdev{85.5}{0.0} \\  %
        \bottomrule
    \end{tabular}
    }
    \caption{ Text results. Evaluation results for the Jargon biomedical tasks ($\uparrow$)}
    \label{tab:jargon}
\end{table*}

\paragraph{Jargon biomedical benchmark}
The Jargon biomedical benchmark \citep[][Table~\ref{tab:jargon}]{segonne-etal-2024-jargon} covers three types of downstream tasks: five token-level classification tasks, one sequence classification task, and one semantic textual similarity task.
The token-level tasks draw on the CAS \citep{grabar2018cas} and ESSAI \citep{dalloux_supervised_2021} corpora for POS tagging and UMLS semantic group prediction (CAS-POS, ESSAI-POS, CAS-SG), the QUAERO FrenchMed corpus \citep{neveol_quaero_2014} for NER on MEDLINE titles and EMEA drug descriptions, and the E3C corpus \citep{minard_european_2021,magnini_e3c_2020} for 3-class BIO entity recognition (E3C-NER), using Layers~1 and~2 for evaluation and fine-tuning respectively.
We report macro-averaged F$_1$ ($\textbf{MacF}_1\uparrow$) for POS tagging and weighted F$_1$ ($\textbf{WF}_1\uparrow$) for the remaining tasks.
FrenchMedMCQA \citep{labrak-etal-2022-frenchmedmcqa} is a multiple-choice QA dataset of 3,105 questions from French medical exams with five answer options each.
CLISTER \citep{hiebel2022clister}, derived from CAS, comprises 1,000 annotated sentence pairs rated 0--5 for semantic similarity in clinical text, evaluated with Spearman's~$\rho$.

\paragraph{Summary results on text tasks}
The results on the text tasks (Tables~\ref{tab:text_eval}--\ref{tab:jargon}) show that \pantaWkMLM{} and \pantaOscMLM{} perform comparably to \camemBWk{} and \flauB{}, while the larger \pantaCrsMLM{} benefits from its extensive pre-training data. On the FLUE benchmark (Table~\ref{tab:text_eval}), \panta{} models lag slightly behind CamemBERT on NLI but match performance on sentiment analysis and Paraphrase tasks, with \pantaOscMLM{} and \pantaCrsMLM{} achieving the best overall balance across tasks, respectively. 

Weaker results on VSD and dependency parsing (Table~\ref{tab:parsing}) suggest that the feature-space objective of data2vec may yield less effective token-level representations for text, even when using the compound loss integrating the MLM objective function.
Conversely, semantic and retrieval-oriented tasks such as classification and extractive QA benefit from large-scale pre-training with \pantaCrsII{} achieving similar results to \camemBOsc{} on the PIAF~1.2 dataset. On the Jargon biomedical benchmark (Table~\ref{tab:jargon}), \panta{} models match or outperform the baselines, except on the \textbf{CLISTER} dataset where the best result has still a significant gap behind the \camemBOsc{} baseline.

Overall, models results and analyses suggest that further improvements can be achieved with a better fine-tuning of the balance between data2vec and MLM loss components, and by employing larger datasets such as OSCAR and CroissantLLM in combination with our best training settings. %
Investigating more effective ways to combine embedding-level and input-level objectives for text models is left for future work.

\subsection{Speech models evaluation}\label{sec:speech-models-evaluation}

\begin{table*}[ht]
    \centering
\resizebox{\textwidth}{!}{
\begin{tabular}{l|cccc|ccc}
\toprule
\multirow{2}{*}{\textbf{Model} }& \textbf{PxCorpus} & \textbf{MEDIA} & \textbf{ETAPE}  & \textbf{AlloSat} & \textbf{fr-en} & \textbf{fr-es} & \textbf{fr-pt} \\
  & NER (F1$\uparrow$) & SLU (CER$\downarrow$) & NER (NEER$\downarrow$) & SER (CCC$\uparrow$) & \multicolumn{3}{c}{ST (BLEU$\uparrow$)}\\ %
\midrule
\wvB & \meanpmstdev{57.9}{3.0} & \meanpmstdev{17.9}{2.0} &  \meanpmstdev{65.13}{0.2} & \meanpmstdev{0.57}{0.10} & \meanpmstdev{14.0}{0.5} & \meanpmstdev{13.2}{0.4} & \meanpmstdev{8.60}{0.3}\\
\wvL & \meanpmstdev{59.9}{3.3} & \meanpmstdev{13.3}{0.6} & \meanpmstdev{52.98}{0.1} &  \meanpmstdev{0.65}{0.02} & \meanpmstdev{18.2}{0.5} & \meanpmstdev{18.2}{0.7} & \meanpmstdev{13.4}{0.4}\\
\wvLL & \meanpmstdev{82.4}{0.2} & \meanpmstdev{12.6}{0.6} & \meanpmstdev{55.30}{0.1}   & \meanpmstdev{0.49}{0.01} & \meanpmstdev{23.1}{0.6} & \meanpmstdev{24.2}{0.6} & \meanpmstdev{21.8}{0.6} \\ \hline
\pantaspeechB & \meanpmstdev{81.7}{1.3} & \meanpmstdev{14.4}{0.7} & \meanpmstdev{61.35}{0.3}  & \meanpmstdev{\textbf{0.84}}{0.02} & \meanpmstdev{17.5}{0.4} & \meanpmstdev{19.0}{0.4} & \meanpmstdev{16.8}{0.3} \\
\pantaspeechL & \meanpmstdev{\textbf{84.4}}{0.4} & \meanpmstdev{\textbf{12.2}}{0.7} & \meanpmstdev{48.14}{0.2} & \meanpmstdev{0.82}{0.02} & \meanpmstdev{24.0}{0.4} & \meanpmstdev{\textbf{25.5}}{0.4} & \meanpmstdev{21.9}{0.4} \\
\pantaspeechLL & \meanpmstdev{\textbf{84.4}}{0.6} & \meanpmstdev{12.3}{0.6} & \meanpmstdev{\textbf{44.68}}{0.4} & \meanpmstdev{0.83}{0.03} & \meanpmstdev{\textbf{25.2}}{0.4} & \meanpmstdev{25.4}{0.4} & \meanpmstdev{\textbf{24.5}}{0.5} \\
\bottomrule
\end{tabular}
}
\caption{ Speech results. Metrics: CER = Concept Error Rate; NEER = Named Entity Error Rate; CCC = Concordance Correlation Coefficient} 
    \label{tab:speech_eval}
\end{table*}

\paragraph{Automatic Speech Recognition (ASR)}
We assess SSL models on ASR in two settings. In the high-resource setting, models are trained on five audiovisual corpora totalling 258 hours: Antract~\cite{carrive_antract}, QUAERO~\cite{boudahmane-etal-2011-advances}, EPAC~\cite{esteve-etal-2010-epac}, ESTER1~\cite{gravier-etal-2004-ester}, and REPERE~\cite{giraudel-etal-2012-repere}.
In the dataset-specific setting, we fine-tune separate models on French CommonVoice version 6.1 \citep{evain:hal-03407172,parcollet:hal-04441389} and ETAPE \citep{gravier2012etape} (low-resource audio-visual).
We also evaluate adaptation to child speech using data from the DyLNet project \citep{nardy2021variation}, comprising 31 hours of fine-tuning data and 4.09/4.16 hours validation/test sets of conversational speech from children aged 3--6.
ASR results in Word Error Rate (WER $\downarrow$) are reported in Table~\ref{tab:asr-results}.

\begin{table*}[ht!]
    \centering
    \resizebox{1.0\linewidth}{!}{
\begin{tabular}{l|cccc|ccc}
\toprule
& \multicolumn{4}{c|}{\textbf{High-resource setting}} & \multicolumn{3}{c}{\textbf{Dataset-specific settings}} \\
\cmidrule(lr){2-5} \cmidrule(lr){6-8}
\textbf{Model} & \textbf{Antract} & \textbf{QUAERO} & \textbf{ESTER1} & \textbf{REPERE} & \textbf{CV 6.1} & \textbf{ETAPE} & \textbf{DyLNet} \\
\midrule
\wvB           & 22.3 \textcolor{gray}{\small$\pm$3.7} & 32.0 \textcolor{gray}{\small$\pm$3.5} & 26.7 \textcolor{gray}{\small$\pm$3.4} & 28.3
\textcolor{gray}{\small$\pm$3.4} & \meanpmstdev{12.28}{0.1} & 34.76 \textcolor{gray}{\small$\pm$0.1} & \meanpmstdev{55.74}{3.7} \\
\wvL           & 9.3 \textcolor{gray}{\small$\pm$0.1}  & 14.7 \textcolor{gray}{\small$\pm$0.2} & 11.4 \textcolor{gray}{\small$\pm$0.1} & 13.5
\textcolor{gray}{\small$\pm$0.1} & \meanpmstdev{9.2}{0.1}& 22.83 \textcolor{gray}{\small$\pm$0.1} & \meanpmstdev{41.89}{0.1} \\
\wvLL          & 8.6 \textcolor{gray}{\small$\pm$0.1}  & 15.0 \textcolor{gray}{\small$\pm$0.3} & 11.5 \textcolor{gray}{\small$\pm$0.2} & 13.6  
\textcolor{gray}{\small$\pm$0.2} & %
\meanpmstdev{9.0}{0.1} 
& 26.03 \textcolor{gray}{\small$\pm$0.1} & \meanpmstdev{41.23}{0.2} \\ \midrule
\pantaspeechB  & 13.2 \textcolor{gray}{\small$\pm$0.3} & 21.3 \textcolor{gray}{\small$\pm$0.4} & 17.9 \textcolor{gray}{\small$\pm$0.4} & 19.1
 \textcolor{gray}{\small$\pm$0.3} & \meanpmstdev{10.5}{0.1} & 30.61 \textcolor{gray}{\small$\pm$0.3} & \meanpmstdev{49.71}{0.3}\\
\pantaspeechL  & 7.5 \textcolor{gray}{\small$\pm$0.1}  & 13.7 \textcolor{gray}{\small$\pm$0.7} & 9.9 \textcolor{gray}{\small$\pm$0.2}  & 11.3
 \textcolor{gray}{\small$\pm$0.2}  & \meanpmstdev{\textbf{8.1}}{0.1}& 19.77 \textcolor{gray}{\small$\pm$0.1} & \meanpmstdev{39.34}{0.2}\\
\pantaspeechLL & \textbf{7.4} \textcolor{gray}{\small$\pm$0.1} & \textbf{11.7} \textcolor{gray}{\small$\pm$0.1} & \textbf{9.7} \textcolor{gray}{\small$\pm$0.1} & \textbf{10.4}
 \textcolor{gray}{\small$\pm$0.1} & \meanpmstdev{8.2}{0.1}& \textbf{19.09} \textcolor{gray}{\small$\pm$0.2} & \meanpmstdev{\textbf{37.48}}{0.2}\\
\bottomrule
\end{tabular}
}
\caption{ Speech results for the ASR tasks in Word Error Rate (WER $\downarrow$) comparing performance on the high-resource setting (columns 2-5) and dataset specific ones (columns 6-8).}
\label{tab:asr-results}
\end{table*}

\paragraph{Named Entity Recognition (NER)}
\label{sec:ner-ETAPE}
We benchmark our models on the French ETAPE corpus~\cite{gravier2012etape}, a 30-hour collection of TV and radio broadcasts covering diverse topics and speaking styles, with an emphasis on spontaneous and multi-speaker speech. ETAPE is a standard benchmark for French NER~\cite{mdhaffar2022end}. We train end-to-end NER systems combining an SSL encoder with a three-layer linear probe. The PxCorpus dataset~\cite{PxCorpus} is also used for speech NER, with labels assigned directly to words or word groups, without using the IOB scheme.

\paragraph{Speech Emotion Recognition (SER)}
\label{sec:ser-allosat}
In this experiment, we focus on SER continuous prediction of affective dimensions (\textit{e.g.}, satisfaction, valence, arousal). We use AlloSat~\citep{macary2020allosat}, a 37-hour corpus of 303 spontaneous French telephone conversations annotated every 250~ms on a frustration $\rightarrow$ satisfaction axis.

\paragraph{Spoken Language Understanding (SLU)}
\label{subsec:slu}
The French MEDIA corpus \citep{bonneau-maynard-etal-2006-results} contains 1\,250 human–machine dialogues about hotel reservations in France, annotated with 76 semantic concepts. It has been widely used to benchmark French SLU systems in both pipeline \citep{Quarteroni.etAl:Interspeech09,dinarelli09:eacl,dinarelli2011:emnlp,dinarelli:hal-01553830,caubriere:hal-02465899,ghannay:hal-03372494} and end-to-end \citep{dinarelli09:Interspeech,dinarelli-etal-2009-ranking,Dinarelli2010:sds,10.1109/ICASSP.2018.8461785,caubriere:hal-02304597,pelloin:hal-03128163,evain:hal-03407172,parcollet:hal-04441389} settings. In particular our results are directly comparable to \citep{evain:hal-03407172,parcollet:hal-04441389}. We perform concept extraction from both speech and transcriptions using Transformer-based SLU models with SSL encoders, reporting Concept Error Rate (CER $\downarrow$) in Tables~\ref{tab:text_eval} and \ref{tab:speech_eval}.

\paragraph{Speech-to-text translation (ST)}
\label{subsec:speech_translation}
The ST task consists in translating speech in a source language into text in another language. We evaluate our speech models on the French-source subsets of the multilingual TEDx corpus~\cite{salesky2021multilingual}, covering three translation directions from French (fr) to English (en), Portuguese (pt), and Spanish (es), with training sets of 50 hours, 38 hours, and 25 hours, respectively. Our experiments are performed in the end-to-end finetuning scenario, where we plug in a 6-layer Transformer decoder to our pre-trained speech SSL models and finetuned the system on each pair. We followed the same settings for ST fine-tuning in Lebenchmark~2.0 paper~\cite{parcollet:hal-04441389} to enable fair comparison with the French wav2vec~2.0-based counterparts. The translation performance in BLEU score~\cite{papineni2002bleu,post2018call} is shown in Table~\ref{tab:speech_eval}. For the LeBenchmark models, we use the results reported in their paper~\cite{parcollet:hal-04441389}.

\paragraph{Summary of results on speech tasks}
Across speech tasks (Tables~\ref{tab:speech_eval} and \ref{tab:asr-results}), \project{} models consistently outperform LeBenchmark baselines. The large model \pantaspeechLL{} achieves the best overall results, with clear gains on challenging spontaneous or noisy corpora such as ETAPE and DyLNet, while maintaining strong performance on CommonVoice. For the cross-modal cross-lingual ST task, the \pantaspeechB model outperforms the \wvB model trained using the same pre-training data by a large margin ($+$5.8 BLEU on average), highlighting the effectiveness of the latent-based objective for speech inputs under medium-resource regimes. Trained on both LeBenchmark and the \inaCentK{} broadcast corpus, it appears particularly robust to acoustic variability. %

Regarding task-specific trends, ASR and SLU benefits most from the larger and more diverse pre-training data, with \pantaspeechLL{} outperforming all other models and \pantaspeechL{} remaining close behind.
In SER, \project{} encoders again excel, with \pantaspeechB{} and \pantaspeechLL{} leading, whereas \wvLL{} struggles to generalize to telephone and spontaneous speech. In the low-resource ST setting (25 hours of fine-tuning data for fr–pt), \pantaspeechLL{} improves translation performance over \pantaspeechL{} by $+$2.6 BLEU, demonstrating the benefits of larger pre-training datasets. Overall, the \project{} family of models demonstrates solid robustness across all speech benchmarks, confirming the benefit of large-scale feature-space pre-training and the diversity brought by the \inaCentK{} corpus.

\section{Discussion and Conclusion}

Across modalities, \project models exhibit complementary behaviors. On speech tasks, \project consistently outperform LeBenchmark baselines, confirming that feature-space prediction is particularly effective for continuous acoustic signals. %
The inclusion of \inaCentK{}, with its large diversity of broadcast and spontaneous conditions, enhances robustness to noise and variability, yielding strong gains on ETAPE and DyLNet, although it slightly reduces performance on cleaner, read-speech data such as CommonVoice. On text tasks, the hybrid {MLM+data2vec} objective mitigates some of the weaknesses of purely embedding-based encoders, achieving competitive results on semantic tasks (QA, classification) but lower scores on syntax-sensitive evaluations (VSD, dependency parsing). This suggests that token-level supervision remains valuable for capturing fine-grained linguistic structure.

Overall, \project models demonstrate that feature-space self-supervision scales efficiently to French.
Its unified architecture, trained separately but identically for text and speech, offers a practical foundation for cross-modal modeling.
To our knowledge, this is the most resource-intensive purely French SSL encoders to date, pairing tens of billions of French tokens with 100k hours of diverse French audio.

Future work will focus on (i) joint optimization on unaligned or weakly aligned speech–text pairs, i.e. multi-modal training. We note that our system has already been modified with respect to data2vec~2.0 and it can be trained on audio and text jointly in multi-modal setting, both aligned and unaligned data. Experiments are in progress; (ii) scaling model capacity and corpus diversity; (iii) extending evaluation to multimodal downstream tasks such as spoken QA or automatic subtitling. Beyond empirical performance, the release of \project aims to empower the French research community with transparent, reproducible, and ethically curated resources, paving the way for responsible multimodal AI in French.

\section{Acknowledgements}
This research has been partially funded by the French National Research Agency (ANR), project "PANTAGRUEL", ANR-23-IAS1-0001.
This work was also supported by the CREMA project (Coreference REsolution into MAchine translation) funded by ANR, contract number ANR-21-CE23-0021-01.
It also received government funding managed by ANR under France 2030, reference ANR-23-IACL-0006.Implementation
It was also supported by ANR through the MIAI "AI \& Language" chair (ANR-19-P3IA-0003) and the MIAI "Socialization and Language at School" chair (ANR-23-IACL-0006). This work was performed using HPC resources from GENCI at IDRIS and CINES under the allocations 2022-A0131013801, 2023-A0151013801, 2024-A0171013801, 2024-A0161015074, and 2025-A0191013801 on the Jean Zay and Adastra supercomputers.

\section*{Bibliographical References}\label{sec:reference}

\bibliographystyle{lrec2026-natbib}
\bibliography{references}

\section*{Language Resource References}
\label{lr:ref}
\bibliographystylelanguageresource{lrec2026-natbib}
\bibliographylanguageresource{languageresource}

\appendix
\section{Pretraining Hyperparameters}\label{sec:appendix-pretraining-hyperparams}
Pre-training hyperparameters for the speech and text models are reported in Tables~\ref{tab:speech-pretraining-hyperparams} and~\ref{tab:text-pretraining-hyperparams}, respectively. All models are trained with the Adam optimizer~\cite{kingma2015adam} ($\beta_1 = 0.9, \beta_2 = 0.98$), using a cosine learning rate scheduler and a weight decay of 0.1.

For text models, the maximum sequence length is set to 512 tokens. We applied the the \emph{complete} break mode in the {fairseq} library, which splits samples only at sentence boundaries but may include multiple sentences per sample. Layer drop is disabled for all text models.

\begin{table*}[!htb]
    \centering
    \begin{tabular}{l| l | l | l}
    \toprule
         & B-1k / B-14k & L-14k & L-114K \\
    \midrule
    Number GPUs & 16 & 48 & 128 \\
    GPU type & H100 & MI250x & H100 \\
    Batch size (seconds/GPU) & 62.5 & 40 & 62.5 \\
    Learning rate & $7.5 \times 10^{-4}$ &$4.0 \times 10^{-4}$  & $2.0 \times 10^{-4}$ \\
    $N_{\textrm{warmup}}$ steps & 8,000 & 5,000 & 20,000\\
    $N_{\textrm{max}}$ & 400,000 & 300,000 & 1,000,000 \\
    Clip norm & - & 1.0 & 1.0 \\
    Layerdrop & 0.05 & 0.0 & 0.0 \\
    Multi-masks & 8 & 12 & 12 \\
    EMA start & 0.999 & 0.9997 & 0.9997 \\
    EMA end & 0.99999 & 1.0 & 1.0 \\
    EMA anneal steps & 75,000 & 300,000 & 600,000 \\
    Mask length & 5 & 5 & 5 \\
    Mask ratio & 0.5 & 0.55 & 0.55\\
    Mask adjust & 0.05 & 0.1 & 0.1 \\
    Top $K$ target layers & 8 & 16 & 16 \\
    Decoder layers & 4 & 4 & 4 \\
    Decoder dimension & 384 & 768& 768\\
    Decoder CNN groups & 16 & 16 & 16 \\
    Decoder kernel & 7 & 7& 7\\
    \bottomrule
    \end{tabular}
    \caption{Pre-training hyper-parameters for speech models.}
    \label{tab:speech-pretraining-hyperparams}
\end{table*}

\begin{table*}[!htb]
    \centering
    \begin{tabular}{l| l | l | l | l}
    \toprule
         & B-camtok-Wk / B-Wk & B-Wk-MLM & B-Osc-MLM & B-Crs-MLM \\
    \midrule
    Number GPUs & 16 & 16 & 32 & 48 \\
    Batch size (tokens/GPU) & 4 & 32 & 28& 28 \\
    Learning rate & $2 \times 10^{-4}$ & $5 \times 10^{-4}$ & $5 \times 10^{-4}$ & $5 \times 10^{-4}$ \\
    $N_{\textrm{warmup}}$ steps & 4,000 & 8,000 & 16,000 & 16,000 \\
    $N_{\textrm{max}}$ & 500,000 & 250,000 & 400,000 & 400,000 \\ 
    
    $\lambda_{\textrm{start}}$ & - & 20.0 & 20.0 & 20.0 \\
    $\lambda_{\textrm{end}}$ & - & 1.0 & 2.0 & 2.0 \\
    $N_{\lambda}$ & - & 250,000 & 400,000 &  400,000 \\
    
    Clip norm & 1.0 & 1.0 & 1.0 & 1.0 \\
    
    Multi-masks & 8 & 8 & 8 & 8\\
    
    EMA start & 0.9999 & 0.9995 & 0.9995&  0.9995 \\
    EMA end & 0.99999 & 0.99995 & 0.99995 & 0.99995\\
    EMA anneal steps & 100,000 & 125,000& 200,000 & 200,000 \\
    
    Mask length & 3 & 3& 3& 3\\
    Mask ratio & 0.6 & 0.6& 0.6& 0.6\\
    Mask adjust & 0.0 & 0.0& 0.0& 0.0\\
    
    Top $K$ target layers & 12 & 12& 12 & 12\\
    
    Decoder layers & 5 & 5& 5& 5\\
    Decoder dimension & 768 & 768& 768& 768\\
    Decoder CNN groups & 1 & 1& 1& 1 \\
    Decoder kernel & 9 & 9& 9 & 9 \\
    \bottomrule
    \end{tabular}
    \caption{Pre-training hyper-parameters for text models. All text models are trained using Nvidia H100 GPU.}
    \label{tab:text-pretraining-hyperparams}
\end{table*}

\section{Details on Benchmark experiments}
\label{sec:appendix_bench}
This section presents text and speech downstream tasks used for evaluation. Unless otherwise noted, all results are averaged over five runs with different random seeds. We use $\uparrow$ (respectively $\downarrow$) to indicate that higher (respectively lower) values correspond to better performance.

\subsection{Text models evaluation}\label{sec:appendix_text-models-eval}
\paragraph{Named Entity Recognition (NER)}
We perform NER on PxCorpus \cite{PxCorpus}, a dataset containing 1\,981 French medical prescriptions in the form of speech recordings along with their transcriptions. Each word of the prescription is tagged in the IOB (Inside-Outside-Beginning) format in various chunks such as ``drug name'', ``frequency'', ``dosage'', \etc. In the case of the text modality, the task involves classifying tokens, aligning them with the corresponding words, and assigning an IOB label to each word by taking the label of the first token of the word. 
    
A linear layer followed by a softmax are added on top of the encoder to predict one of the 37 different classes for each token. The encoder and the linear layer are fine-tuned together during training. The loss function used is the cross-entropy loss weighted to account for class imbalance.
The metric used for evaluating the models is the F1-score from the SeqEval library \cite{seqeval} which is designed for IOB-tagging evaluation.

\paragraph{Automatic Coreference Resolution (CR)}
The task consists of identifying all mentions (\ie, phrases referring to entities) in the text and grouping those that co-refer to the same entity into clusters. Due to the inherent complexity of coreference resolution, recent systems adopt diverse approaches, often tailored to specific datasets or domains. 
For benchmarking, we selected the WL-Coref model \citep{dobrovolskii-2021-word,doosterlinck-etal-2023-caw,liu-etal-2024-mscaw}, known for its strong performance and computational efficiency during training and evaluation. All experiments were implemented using the Stanza library \citep{qi2020stanza}. 

For evaluation, we used ANCOR, a corpus comprising manually transcribed interviews \citep{muzerelle-etal-2014-ancor}. The corpus is freely available as part of CorefUD 1.3, a multilingual collection of coreference corpora \citep{novak-etal-2025-corefud}. Since only train and validation splits are available, we used 10\% of the train split for model validation during training and the validation split for reporting the performance. 
System performance was assessed using the official \texttt{corefud-scorer} \citep{novak-etal-2024-findings}, with head mention matching as the evaluation criterion.
We report the CoNLL F1 score. %

\paragraph{Extractive Question Answering (QA)}
The task consists in finding the shortest span of words in a context answering a given question. We used PIAF1.2 \citep{keraron-EtAl:2020:LREC}, an extractive question answering dataset for French inspired by SQuAD1.1 \citep{rajpurkar2016squad100000questionsmachine} which contains 9\,224 context/question/answer pairs extracted from French Wikipedia. 

The context and the question are passed in the encoder separated by a <SEP> special token. A linear layer with two outputs is added on top of the encoder to predict, for each token, the logits that the token starts and ends the answer span. The predicted span is obtained by choosing the tuple $(\mathrm{token1}, \mathrm{token2})$ in the context that maximizes $\mathrm{token1}[\mathrm{start}] + \mathrm{token2}[\mathrm{end}]$ (the sum is used here instead of the product because the linear layer produces logits and not probabilities). At the end, the context span expressed in character numbers and not token numbers is obtained by taking the smallest word span that contains the predicted token span.
The encoder and the linear layer are fine-tuned together during the training phase. The cross-entropy loss is used for computing the gradient.

The metric used for evaluation is a F1-score on words, where words are categorized in True/False Positives/Negatives depending on their inclusion in the predicted and ground-truth word spans. This F1-score is at the word-level and not the token-level as different models use different tokenizers that segment the same text into varying token counts. As a result using a token-level F1-score would lead to different evaluations for different models.

\paragraph{FLUE benchmark} 
We evaluate the text models on part of the tasks included in the FLUE benchmark  \citep{le2020flaubertunsupervisedlanguagemodel}. The tasks are as follows: text classification, paraphrase identification, natural language inference (NLI), verb sense disambiguation, and dependency parsing. The overall methodology and downstream model architecture are based on the FLUE framework described by \citet{le2020flaubertunsupervisedlanguagemodel}. %

For text classification, we use the Cross-Lingual Sentiment (CLS) dataset \citep{prettenhofer:2010}, formulated as a binary classification task that aims to determine whether a given review expresses a positive or negative sentiment. For paraphrase identification, we use the Cross-Lingual Adversarial Dataset for Paraphrase Identification \citep[PAWS-X,][]{zhang2019pawsparaphraseadversariesword}, also a binary classification task where the goal is to assess whether two input sentences are semantically equivalent. The reported results include the macro F1 score and their standard deviation across five different seeds. For natural language inference , we use the XNLI dataset \citep{conneau-etal-2018-xnli}, a three-class classification task that determines the logical relationship between a premise and a hypothesis. Following \citet{martin-etal-2020-camembert} and \citet{le2020flaubertunsupervisedlanguagemodel}, we report the mean and standard deviation from 10 runs with different seeds. The output label indicates whether the hypothesis is entailed by, contradicts, or is neutral with respect to the premise.  

For verb sense disambiguation, we evaluate on the FrenchSemEval dataset \citep{segonne-etal-2019-using}, which tests sense disambiguation of French verbs.  We report the macro F1 score as the evaluation metric.

For dependency parsing, we use a set of treebanks from the Universal Dependencies repository \citep{de-marneffe-etal-2021-universal}: Sequoia \citep{candito-etal-2014-deep}, GSD \citep{mcdonald-etal-2013-universal}, as well as two treebanks for spoken French: Rhapsodie \citep{lacheret-etal-2014-rhapsodie} and ParisStories \cite{kahane-etal-2021-annotation}. %
We use the HOPS parser \citep{grobol:hal-03223424} and report classic evaluation metrics for the task: Labelled Attachment Score (LAS) and part-of-speech accuracy (POS).

\paragraph{Jargon biomedical benchmark}
The Jargon biomedical benchmark \citep[][Table~\ref{tab:jargon}]{segonne-etal-2024-jargon} covers three types of downstream tasks.
Five of the seven tasks are token-level classification tasks, which are accompanied by a sequence classification task and a semantic textual similarity task.

The token-level labeling tasks are built on CAS \citep{grabar2018cas} and ESSAI corpus \citep{dalloux_supervised_2021}, used for POS tagging and UMLS semantic group prediction (CAS-POS, ESSAI-POS, CAS-SG).
The QUAERO FrenchMed corpus \citep{neveol_quaero_2014} provides NER data from MEDLINE titles and EMEA drug descriptions.
We also employ the E3C corpus \citep{minard_european_2021,magnini_e3c_2020} for a 3-class BIO entity recognition task (E3C-NER), using Layers 1 (manual annotation) and Layer 2 (automatic labeling) for evaluation and fine-tuning respectively. %
We report macro-averaged F$_1$ scores for the POS-tagging tasks ($\textbf{MacF}_1\uparrow$); for the others, a weighted average across the output classes ($\textbf{WF}_1\uparrow$).

FrenchMedMCQA \citep{labrak-etal-2022-frenchmedmcqa} is a multiple-choice QA dataset of 3,105 questions from French medical specialization exams, each with five answer options.
Since questions can have multiple correct answers, the Hamming score we use measure the overlap between the predicted answer combination and the ground truth.
CLISTER \citep{hiebel2022clister}, derived from CAS, comprises 1,000 manually annotated sentence pairs (scores 0–5) and is used to assess semantic similarity in clinical text, a task important for detecting redundant information \citep{wang2020medsts}.
We use Spearman's correlation coefficient~$\rho$ for evaluation. 

\subsection{Speech models evaluation}\label{sec:appendix_speech-models-eval}

\paragraph{Automatic Speech Recognition (ASR)}
We assess SSL models on ASR in two settings: with a high-resource pipeline  where we train on multiple corpora, and dataset specific pipelines. For each pipeline, we use a character-based CTC architecture with an optimized number of linear layers and learning rate. Linear layers are placed on top of the SSL encoder, and the whole model is finetuned. Transcriptions are obtained with a greedy decoding and without the use of a language model.

High-resource setting use multiple audiovisuals training datasets, representing 258h of training data: Antract~\cite{carrive_antract},
QUAERO~\cite{boudahmane-etal-2011-advances},
EPAC~\cite{esteve-etal-2010-epac},
ESTER1~\cite{gravier-etal-2004-ester} and
REPERE~\cite{giraudel-etal-2012-repere}.
    
Next, we assess SSL models on ASR on a dataset specific settings, where we train specific variations for each dataset with similar architectures.
We first benchmark models on the French CommonVoice (CV) dataset. We use the CommonVoice 6.1 version to compare with results from \citet{evain:hal-03407172,parcollet:hal-04441389}.
Similarly, we use ETAPE \citep{gravier2012etape} dataset for ASR on low-resource audiovisual settings.
    
Finally, we assessed the ability of ASR models to adapt to spontaneous child speech. For this purpose, we used data from the DyLNet project \citep{nardy2021variation}, which ecologically collected conversational speech from children aged 3 to 6 years in a school context. The resulting corpus comprises approximately 35\,000 hours of raw audio, of which 800 hours were manually annotated. Among these annotations, 60 hours correspond to speech segments. From this subset, 31 hours were selected for fine-tuning the SSL models. The validation and test sets consist of 4.09 and 4.16 hours of speech, respectively.    
We present in Table \ref{tab:asr-results} the ASR results in Word Error Rate (WER $\downarrow$) for those pipelines. %

\paragraph{Named Entity Recognition (NER)}
\label{sec:appendix-ner-ETAPE}
We benchmark our models on the French ETAPE corpus~\cite{gravier2012etape}, a 30-hour collection of TV and radio broadcasts covering diverse topics and speaking styles, with an emphasis on spontaneous and multi-speaker speech. ETAPE is a standard benchmark for French NER~\cite{mdhaffar2022end}. We train end-to-end NER systems combining an SSL encoder with a three-layer linear probe. The PxCorpus dataset~\cite{PxCorpus} is also used for speech NER, with labels assigned directly to words or word groups, without the IOB scheme.

\paragraph{Speech Emotion Recognition (SER)}
\label{sec:appendix-ser-allosat}
\textit{Speech Emotion Recognition (SER)} comprises (1) discrete-state single-label classification (such as happy, sad, or neutral) and (2) continuous estimation of a given affective dimension (\textit{e.g.} satisfaction, valence, arousal), yielding an intensity time series. We focus on continuous prediction.
We use AlloSat~\citep{macary2020allosat}, a corpus of 303 spontaneous French telephone conversations (total 37h), with time-continuous annotations on a frustration$\rightarrow$satisfaction axis (a label every 250~ms).

\subparagraph{Features and alignment} We use frozen self-supervised encoders to extract frame-level embeddings produced approximately every 20~ms, while labels are provided at the annotation rate. To align these rates, we aggregate consecutive frames into 250~ms windows by alternating mean-pooling over 12 and 13 frames (average 12.5 $\approx$ 250~ms), so the feature and label timelines stay synchronized over long sequences. We drop the final partial window, compute the train-set mean and standard deviation per feature dimension, and apply this normalization to all splits.

\subparagraph{Model and hyperparameters}\quad
The downstream model is a 5-BiLSTM (five stacked bidirectional LSTM layers with hidden sizes $[512, 200, 64, 32, 32]$) followed by a linear layer predicting one scalar per frame. Training minimizes negative CCC between predictions and labels using Adam (lr $=1\times10^{-3}$), batch size 15, for 200 epochs. For development and test, we compute a single concatenated CCC per split over all valid %
. We then select the checkpoint with the best dev-CCC (best over epochs) and evaluate that checkpoint on the test set.

\subparagraph{Results: analysis and discussion for SER}\quad
Performances vary across LeBenchmark models (3K/7K/14K), while \project encoders are consistently strong in this frozen regression setup. \pantaspeechB gets the best results on both the validation and test data; \pantaspeechLL is close on test. 
In contrast, LeBenchmark-14K-large is weak on test, indicating limited off-the-shelf generalization to telephone conversations. 
These findings suggest that, in this frozen-encoder regression setting, more pretraining hours or larger model size do not necessarily yield better SER performance.

\paragraph{Spoken Language Understanding (SLU)}
\label{subsec:appendix-slu}

We benchmark our models on the SLU task with the French corpus MEDIA \citep{bonneau-maynard-etal-2006-results}. This corpus covers the topic of hotel information and reservations in France, and is made up of 1,250 human-machine dialogues transcribed and annotated with 76 semantic concepts. \\
MEDIA has been used extensively for benchmarking French models for SLU, both with symbolic and neural models, and both in pipeline systems, where an Automatic Speech Recognizer (ASR) feeds a Natural Language Understanding (NLU) module \citep{Quarteroni.etAl:Interspeech09,dinarelli:hal-01553830}, and \emph{end-to-end} systems \citep{dinarelli09:Interspeech,Dinarelli2010:sds}. In particular we can compare results presented here with those published in \citep{evain:hal-03407172,parcollet:hal-04441389}.
Downstream SLU models are based on the Transformer architecture and use either speech or text SSL encoders. All models are trained with a 3-stage learning rate (lr) scheduler, with 25 epochs of linear lr warm-up to reach the optimal learning rate, 65 epochs during which lr is unchanged, and 60 epochs where lr decays exponentially. The optimal lr for all text models and \textit{Base} speech models is $5e^{-5}$, while we use $2.5e^{-5}$ for \textit{Large} speech models.
Like in \citep{parcollet:hal-04441389} we compute the average of the 5 best models based on the loss on the development data and we perform  greedy decoding. The evaluation metric is the Concept Error Rate (the lower the better $\downarrow$), and we report the average of 5 runs.

Results are reported in Tables~\ref{tab:text_eval} and \ref{tab:speech_eval} as \textit{SLU} in terms of Concept Error Rate (CER $\downarrow$).

We note that when performing SLU from speech, and in order to extract both concepts (slots) and values (slot-values), the model predicts words, concept boundaries and concepts at the same time. For instance from a speech segment containing the text segment \emph{``... in Majorca from the 11th to the 15th of May ...''}, the model should predict \emph{``<Begin> in Majorca LOCALIZATION <End> <Begin> from the 11th START-DATE <End> <Begin> to the 15th of May END-DATA <End>''}. 
This output is then post-processed as \emph{``LOCALIZATION[in Majorca] START-DATE[from the 11th] END-DATE[to the 15th of May]''}.
Finally, the value extraction phase is performed, and the final output is \emph{``LOCALIZATION[Majorca] START-DATE[11/05/2026] END-DATE[15/05/2026]''}.

Because of the presence of open-value concepts, such like dates and other amounts, value extraction has been performed for long with rule-based systems \cite{Hahn.etAL:SLUJournal:2010}. However this has been recently questioned since rules appeared to have a high oracle error rate (i.e. error rate over gold reference concepts) \cite{laperriere-etal-2022-spoken}.
Thus \cite{mdhaffar2025sensemodelsopensource} evaluated SLU models with Concept Error Rate (CER in this work), that is error rate over sequences of concepts alone (i.e. \emph{``LOCALIZATION START-DATE END-DATA''} in the example above), and on Concept-Value Error Rate (CVER) where values are actually unnormalized phrases with tokens concatenated to each other (i.e. \emph{``LOCALIZATION[in-Majorca] START-DATE[from-the-11th] END-DATE[to-the-15th-of-May]''}).

While we are able to extract both concepts and concept-values, to keep this work comparable with to \citep{evain:hal-03407172,parcollet:hal-04441389}, and for lack of space in the tables and in the paper, in this work we show only CER. We leave evaluation with CVER, and possibly more sophisticated approaches for normalized value extraction, to future work.

\begin{table*}[ht]
\centering
\small

\renewcommand{\arraystretch}{1.5}
\begin{tabularx}{\textwidth}{l | l | p{2.8cm} | X | l}
\toprule
\textbf{Category} & \textbf{Task} & \textbf{Datasets} & \textbf{Description} & \textbf{Metrics} \\ \midrule

\multirow{4}{*}{General} & NER & PxCorpus & Medical prescription entity extraction (IOB format). & SeqEval F1 $\uparrow$ \\ \cline{2-5}
 & Coreference & ANCOR & Identifying and clustering mentions of the same entity. & CoNLL F1 $\uparrow$ \\ \cline{2-5}
 & Extractive QA & PIAF 1.2 & Finding answer spans within French Wikipedia contexts. & Word-level F1 $\uparrow$ \\ \cline{2-5}
 & SLU & MEDIA & Concept extraction from manually transcribed hotel reservation dialogues. & CER $\downarrow$ \\ \midrule

\multirow{5}{*}{FLUE} & Classification & CLS & Binary sentiment analysis of Amazon reviews. & Macro F1 $\uparrow$ \\ \cline{2-5}
 & Paraphrase & PAWS-X & Semantic equivalence detection between sentence pairs. & Macro F1 $\uparrow$ \\ \cline{2-5}
 & NLI & XNLI & Logical relationship classification (Entailment, etc.). & Accuracy $\uparrow$ \\ \cline{2-5}
 & VSD & FrenchSemEval & Disambiguating French verb meanings in context. & Macro F1 $\uparrow$ \\ \cline{2-5}
 & Dependency & Sequoia, GSD, Rhapsodie, ParisStories & Syntactic dependency parsing and POS tagging. & LAS $\uparrow$ / POS $\uparrow$ \\ \midrule

\multirow{6}{*}{\shortstack[l]{Jargon\\Biomedical}} & POS Tagging & ESSAI-POS, CAS-POS & Token-level part-of-speech tagging for medical text. & Macro F1 $\uparrow$ \\ \cline{2-5}
 & NER & MEDLINE, EMEA & Entity recognition on drug descriptions and titles. & Weighted F1 $\uparrow$ \\ \cline{2-5}
 & Semantic Group. & CAS-SG & Predicting UMLS semantic groups for medical terms. & Weighted F1 $\uparrow$ \\ \cline{2-5}
 & BIO Recognition & E3C & 3-class entity recognition in clinical case reports. & Weighted F1 $\uparrow$ \\ \cline{2-5}
 & Medical QA & FrMedMCQA & Multiple-choice French medical examination questions. & Hamming $\uparrow$ \\ \cline{2-5}
 & Similarity & CLISTER & Semantic similarity (0–5 scale) for clinical sentences. & Spearman's $\rho \uparrow$ \\
\bottomrule
\end{tabularx}

\caption{Overview of French General and Biomedical Text Model Evaluation Benchmarks}
\label{tab:overview_text_benchmarks}

\end{table*}

\begin{table*}[ht!]
\centering
\small
\renewcommand{\arraystretch}{1.5}
\begin{tabularx}{\textwidth}{l | l | p{2.8cm} | X | l}
\toprule
\textbf{Task} & \textbf{Subtask} & \textbf{Datasets} & \textbf{Description} & \textbf{Metrics} \\ \midrule

\multirow{3}{*}{ASR} & High-resource & Antract, QUAERO, ESTER1, REPERE & Broad-scale ASR on 258h of audiovisual data. & WER $\downarrow$ \\ \cline{2-5}
 & Specific & CommonVoice, ETAPE & Benchmarking on crowdsourced and low-resource data. & WER $\downarrow$ \\ \cline{2-5}
 & Child Speech & DyLNet & Conversational speech recognition for children (ages 3–6). & WER $\downarrow$ \\ \midrule

\multirow{2}{*}{Speech NER} & Broadcast & ETAPE & End-to-end entity recognition on TV/Radio broadcasts. & NEER $\downarrow$ \\ \cline{2-5}
 & Medical & PxCorpus & End-to-end entity labeling on medical recordings. (direct to word groups, no IOB schema) & F1 $\uparrow$ \\ \midrule

\multirow{1}{*}{SER} & Continuous & AlloSat & Continuous frustration-satisfaction axis prediction. & CCC $\uparrow$ \\ \midrule

\multirow{1}{*}{SLU} & Concept Ext. & MEDIA & End-to-end spoken language understanding for hotel reservation dialogues. & CER $\downarrow$ \\ \midrule

\multirow{1}{*}{ST} & Translation & mTEDx & Multilingual speech-to-text translation (fr $\to$ en, es, pt). & BLEU $\uparrow$ \\
\bottomrule
\end{tabularx}
\caption{Overview of Speech Model Evaluation Benchmarks}
\label{tab:overview_speech_benchmarks}
\end{table*}

\begin{table*}[!p]

\parbox{\columnwidth}{
  \section{Speech datasets}
  \label{sec:speech_dataset_appendix}
}
\vspace{1em}

\centering 
\scalebox{0.85}{ 
\begin{tabular}{p{5cm}lcl}
\toprule 
\textbf{Corpus} & \textbf{License} & \textbf{Duration} & \textbf{Speech type}\\

\toprule 
\multicolumn{4}{c}{\textbf{1K dataset}} \\ 
\midrule

\makecell[l]{MLS French\\\cite{old_pratap_mls_2020}} & CC BY 4.0 & \begin{tabular}[c]{@{}c@{}}\textbf{1,096:43} \\ 520:13 / 576:29 / --\end{tabular} & Read\\ 
\bottomrule

\multicolumn{4}{c}{\textbf{14k dataset}} \\ 
\midrule

\makecell[l]{EPAC**\\\cite{esteve-etal-2010-epac}} & ELRA NC & \begin{tabular}[c]{@{}c@{}}\textbf{1,626:02} \\ 1,240:10 / 385:52 / --\end{tabular} & \begin{tabular}[c]{@{}l@{}}Radio \\ Broadcasts\end{tabular} \\
\hline

\makecell[l]{African Accented French\\\cite{noauthor_african_2003}} & Apache 2.0 & \begin{tabular}[c]{@{}c@{}}\textbf{18:56} \\ -- / -- / 18:56\end{tabular} & Read\\
\hline

\makecell[l]{Att-Hack\\\cite{le_moine_att-hack_2020}} & CC BY-NC-ND & \begin{tabular}[c]{@{}c@{}}\textbf{27:02} \\ 12:07 / 14:54 / --\end{tabular} & \begin{tabular}[c]{@{}l@{}}Acted \\ Emotional\end{tabular} \\
\hline

\makecell[l]{CaFE\\\cite{gournay_canadian_2018}} & CC NC & \begin{tabular}[c]{@{}c@{}}\textbf{1:09} \\ 0:32 / 0:36 / --\end{tabular} & \begin{tabular}[c]{@{}l@{}}Acted \\ Emotional\end{tabular} \\
\hline

\makecell[l]{CFPP2000*\\\cite{branca-rosoff_discours_2012}} & CC BY-NC-SA & \begin{tabular}[c]{@{}c@{}}\textbf{16:26}\\ 0:14 / 1:56 / 14:16\end{tabular} & Spontaneous\\
\hline

\makecell[l]{ESLO2\\\cite{eshkol-taravella_grand_2011}} & CC BY-NC-SA & \begin{tabular}[c]{@{}c@{}}\textbf{34:12} \\ 17:06 / 16:57 / 0:09 \end{tabular} & Spontaneous \\
\hline

\makecell[l]{GEMEP\\\cite{banziger_introducing_2012}} & User agreement & \begin{tabular}[c]{@{}c@{}}\textbf{0:50} \\ 0:24 / 0:26 / --\end{tabular} & \begin{tabular}[c]{@{}l@{}}Acted \\ Emotional\end{tabular} \\
\hline

\makecell[l]{MPF\\\cite{MPF_paper_2017,MPF_ortolang}} & CC BY-NC-SA 4.0 & \begin{tabular}[c]{@{}c@{}}\textbf{19:06} \\ 5:26 / 4:36 / 9:03\end{tabular} & Spontaneous \\
\hline

\makecell[l]{PORTMEDIA (French)\\\cite{lefevre_robustesse_2012}} & ELRA NC & \begin{tabular}[c]{@{}c@{}}\textbf{38:59}\\ 19:08 / 19:50 / --\end{tabular} & \begin{tabular}[c]{@{}l@{}}Acted telephone \\ dialogue \end{tabular} \\
\hline

\makecell[l]{TCOF (Adults)\\\cite{TCOF_ortolang}} & CC BY-NC-SA & \begin{tabular}[c]{@{}c@{}}\textbf{53:59} \\ 9:33 / 12:39 / 31:46\end{tabular} & Spontaneous \\
\hline

\makecell[l]{NCCFr\\\cite{torreira_nijmegen_2010}} & User agreement & \begin{tabular}[c]{@{}c@{}}\textbf{26:35} \\ 12:44 / 12:59 / 00:50\end{tabular} & Spontaneous \\
\hline

\makecell[l]{Voxpopuli (\textit{Unlabeled})\\\cite{wang2021voxpopuli}} & CC0 & \begin{tabular}[c]{@{}c@{}}\textbf{4,532:17}\\ -- / -- / 4,532:17\end{tabular} & \begin{tabular}[c]{@{}l@{}}Professional \\ speech \end{tabular} \\
\hline

\makecell[l]{Voxpopuli (\textit{Transcribed})\\\cite{wang2021voxpopuli}} & CC0 & \begin{tabular}[c]{@{}c@{}}\textbf{211:57}\\ -- / -- / 211:57\end{tabular} & \begin{tabular}[c]{@{}l@{}}Professional \\ speech \end{tabular} \\
\midrule

\makecell[l]{Audiocite.net\\\cite{felice2024audiocite}} & CC BY + ND/NC/SA & \begin{tabular}[c]{@{}c@{}}\textbf{6698:35}\\ 3477:24 / 1309:49 / 1911:21 \end{tabular} & \begin{tabular}[c]{@{}c@{}} Read \end{tabular} \\
\midrule

\makecell[l]{Niger-Mali Audio collection\\ \cite{zanon-boito-etal-2022-speech}} & CC BY-NC-ND & \begin{tabular}[c]{@{}c@{}}\textbf{111:01}\\ 52:15 / 58:46 / --\end{tabular} & \begin{tabular}[c]{@{}l@{}} Radio \\ broadcasts \end{tabular} \\
\midrule

\textbf{14K dataset total} & & \begin{tabular}[c]{@{}c@{}}\textbf{14,529:18} \\ 5,363:01 / 2,409:42 / 6,756:28 \end{tabular} & - \\

\bottomrule
\multicolumn{4}{c}{\textbf{114k dataset}} \\ 
\midrule

\inaCentK{}** & private & \begin{tabular}[c]{@{}c@{}}\textbf{100,000:00} \\ 52,958:00 / 22,642:00 / --\end{tabular} & \begin{tabular}[c]{@{}l@{}}Radio/TV \\ Broadcasts\end{tabular} \\\midrule

\textbf{114k dataset total} & & \begin{tabular}[c]{@{}c@{}}\textbf{114,529:18} \\ 58,321:01 / 25,051:42 / 6,756:28 \end{tabular} & - \\

\bottomrule 
\end{tabular}
}
\caption{Statistics for the speech corpora used to train SSL models according to gender information (male / female / unknown). The small dataset is from MLS only. Each dataset is composed of the previous one + additional data; duration: hour(s):minute(s).}  
\raggedright \scriptsize{*Composed of audio files not included in the CEFC corpus v2.1, 02/2021; **speakers are not uniquely identified.; Stats of CFPP2000, MPF and TCOF have changed a bit due to a change in data extraction; License: CC=Creative Commons; NC=non-commercial; BY= Attribution; SA= Share Alike; ND = No Derivative works; CC0 = No Rights Reserved; User agreement = open for research and NC.
} \label{tab:speech_datasets_simplified} 
\end{table*}

\end{document}